\documentclass[pdflatex,sn-mathphys-num]{sn-jnl}

\usepackage{graphicx}%
\usepackage{multirow}%
\usepackage{amsmath,amssymb,amsfonts}%
\usepackage{amsthm}%
\usepackage{mathrsfs}%
\usepackage[title]{appendix}%
\usepackage{xcolor}%
\usepackage[normalem]{ulem}%
\usepackage{textcomp}%
\usepackage{manyfoot}%
\usepackage{booktabs}%
\usepackage{algorithm}%
\usepackage{algorithmicx}%
\usepackage{algpseudocode}%
\usepackage{listings}%
\usepackage{svg}%
\usepackage{subcaption}%
\usepackage{titlesec}%
\usepackage{tabularx}%
\usepackage{pdflscape}%
\usepackage{makecell}%
\usepackage{lineno}%
\usepackage{hyperref}
\usepackage{pdflscape}%
\usepackage{verbatim}%
\usepackage{ragged2e}%
\usepackage{array}%

\usepackage[T1]{fontenc}
\usepackage[utf8]{inputenc} 

\newcolumntype{L}{>{\RaggedRight\arraybackslash}X}

\begin{document}

\title{Advanced techniques and applications of LiDAR Place Recognition in Agricultural Environments: A Comprehensive Survey}

\author*[1]{\fnm{Judith} \sur{Vilella-Cantos}}\email{jvilella@umh.es}

\author[1]{\fnm{Mónica} \sur{Ballesta}}\email{m.ballesta@umh.es}

\author[1]{\fnm{David} \sur{Valiente}}\email{d.valiente@umh.es}

\author[1]{\fnm{María} \sur{Flores}}\email{m.flores@umh.es}

\author[1]{\fnm{Luis} \sur{Payá}}\email{lpaya@umh.es}

\affil*[1]{\orgdiv{University Institute for Engineering Research}, \orgname{Miguel Hernández University}, \orgaddress{\street{Avda. de la Universidad s/n, Edificio Innova}, \city{Elche}, \postcode{03202}, \state{Alicante}, \country{Spain}}}


\abstract{An optimal solution to the localization problem is essential for developing autonomous robotic systems. Apart from autonomous vehicles, precision agriculture is one of the fields that can benefit most from these systems. Although LiDAR place recognition is a widely used technique in recent years to achieve accurate localization, it is mostly used in urban settings. However, the lack of distinctive features and the unstructured nature of agricultural environments make place recognition challenging. This work presents a comprehensive review of state-of-the-art the latest deep learning applications for agricultural environments and LPR techniques. We focus on the challenges that arise in these environments. We analyze the existing approaches, datasets, and metrics used to evaluate LPR system performance and discuss the limitations and future directions of research in this field. This is the first survey that focuses on LiDAR-based localization in agricultural settings, with the aim of providing a thorough understanding and fostering further research in this specialized domain.}


\maketitle

\section{Introduction}\label{sec:intro}
The increasing global population has led to a connected rise in the demand for food production. In order to address this growing demand, the agricultural sector is increasingly turning to technological advancements and automation to enhance efficiency and productivity. Additionally, the rising average age of the agricultural workforce is a contributing factor to the demand for automation \cite{shaikh2022recent}. The integration of autonomous robotic systems into agricultural practices could induce a paradigm shift within the field, transforming the landscape of agriculture in the following ways: firstly, the implementation of such systems could lead to a reduction in labor costs; secondly, there is the possibility of increasing crop yields; and thirdly, the environmental impact could be minimized \cite{dhanya2022deep, attri2023review}. However, for these systems to operate effectively, they must possess the capability to accurately localize themselves within their environment.

Place recognition approaches have been extensively utilized in the domain of robotics to achieve precise localization. These techniques entail the robot's ability to recognize previously visited locations based on sensory data, thereby determining its position within a map \cite{masone2021survey, garg2021your}. Light Detection and Ranging (LiDAR) sensors are frequently employed for place recognition due to their capacity to provide precise three-dimensional information about the environment. Furthermore, while vision solutions have historically dominated the field, LiDAR sensors have demonstrated enhanced resilience to variations in lighting and weather conditions \cite{zhang2024lidar}.

The performance of LiDAR place recognition (LPR) is highly dependent on the characteristics of the environment. A significant number of recent approaches emphasize the use of urban datasets for the training of Deep Learning (DL) models \cite{du20243}. However, these models may not always exhibit optimal generalizability when applied to agricultural environments. The presence of characteristic elements, such as buildings, roads, and vehicles, in urban environments can be readily identified by sophisticated algorithms. Conversely, agricultural environments frequently exhibit characteristics that are unstructured and void of distinctive features, thereby compromissing the efficacy of place recognition algorithms in accurately identifying locations \cite{ding2022recent}.

This survey addresses agricultural environments as an emerging field for LPR, providing comprehensive review of the state-of-the-art (SOTA), with a special focus on the challenges that arise in agricultural environments. Prior to that, we delve into the evolution of agricultural applications with DL approaches, with a focus on the recent advances and needs within the localization problem.

The contributions of this work are as follows:
\begin{itemize}
    \item A comprehensive review of the SOTA in LPR, concentrating on actual challenges within agricultural settings.
    \item An analysis of the different existing approaches, datasets and metrics used to evaluate the performance of localization approaches in agricultural scenarios.
    \item A discussion of the limitations and future directions for research in this field.
\end{itemize}

\subsection{Scope of the survey}\label{subsec:scope}
The present survey focuses on the use of LiDAR for place recognition in agricultural environments. While numerous surveys on place recognition in urban environments already exist, there is a lack of comprehensive reviews that specifically address the challenges and solutions for agricultural settings. The objective of this survey is to bridge this knowledge gap by offering a comprehensive analysis of the SOTA in DL applications in agricultural environments, with a particular focus on the LPR problem. In this study, particular emphasis is placed on methodologies employed from 2020 onward, as the field has witnessed substantial advancements in recent years, particularly with the emergence of DL techniques. In addition, relevant works from earlier years have been included to provide context and background for the current SOTA.

\subsection{Organization of the survey}\label{subsec:organization}
The remainder of the survey is structured as follows. Section \ref{sec:agri_DL_applications} explores the main applications of DL in these settings over the past several years. A thorough examination of the existing DL solutions for localization problems, such as place recognition and SLAM, in agricultural environments is presented in Section \ref{sec:existing_solutions_agri}. This section encompasses both vision-based and LiDAR-based methodologies. In addition, this section provides a detailed discussion of the specific challenges that arise in agricultural environments for localization. In Section \ref{sec:LPR_problem} the genesis and significance of the LPR problem are examined through a comprehensive lens. Section \ref{sec:datasets} reviews the existing datasets collected in agricultural settings. These datasets are examined in terms of their traditional use in detection and segmentation. As outlined in Section \ref{sec:metrics}, the evaluation metrics employed to assess the performance of LPR algorithms are discussed. Finally, the concluding Section \ref{sec:conclusions}, presents a thorough summary of the research findings and envisages several key insights for future studies in this field.

\section{Agricultural deep learning applications}\label{sec:agri_DL_applications}
The application of DL in agricultural environments has expanded rapidly in recent years. While traditional computer vision methods often struggle with the variability of outdoor settings, supervised and unsupervised DL models have demonstrated remarkable performance in core perception tasks, such as semantic segmentation, object detection, crop classification, and fruit counting.

However, deploying a localization DL model in real-world agricultural scenarios remains non-trivial compared to structured industrial or urban environments. Agricultural settings are characterized by high unstructuredness, severe occlusions, variable lighting conditions, and seasonal appearance changes \cite{barros2024spvsoap3d, marzoa2024magro}. These factors demand robust architectures capable of generalizing across different crop stages and environmental conditions.

In this section, we review the primary applications of DL in agriculture, categorizing them by their specific utility. We analyze how SOTA approaches address the unique unstructured nature of these environments to achieve reliable perception.


\subsection{Fields of application}\label{subsec:applications_agri}
According to the European Commission, the average age of farmers in the European Union (EU) is 57 years, with only 12\% of farmers under the age of 40. This suggests that the agricultural sector is experiencing a shortage of labor \cite{eu2025farmers}. The integration of robotics within agricultural settings is driven by the objective of enhancing efficiency, mitigating labor expenses, and augmenting crop yields \cite{cheng2023recent}. These agricultural robots have the capacity to execute tasks such as planting, harvesting, and crop monitoring with a high degree of precision. To do so, a reliable sensor suite is imperative for the successful operation of these robotic systems.  In this section, we will examine several primary applications of DL techniques in agricultural environments, with a particular emphasis on the past five years.

\subsubsection{Precision agriculture}\label{subsubsec:precision_agri}
\begin{figure}[h]
    \centering
    \includegraphics[width=0.8\textwidth]{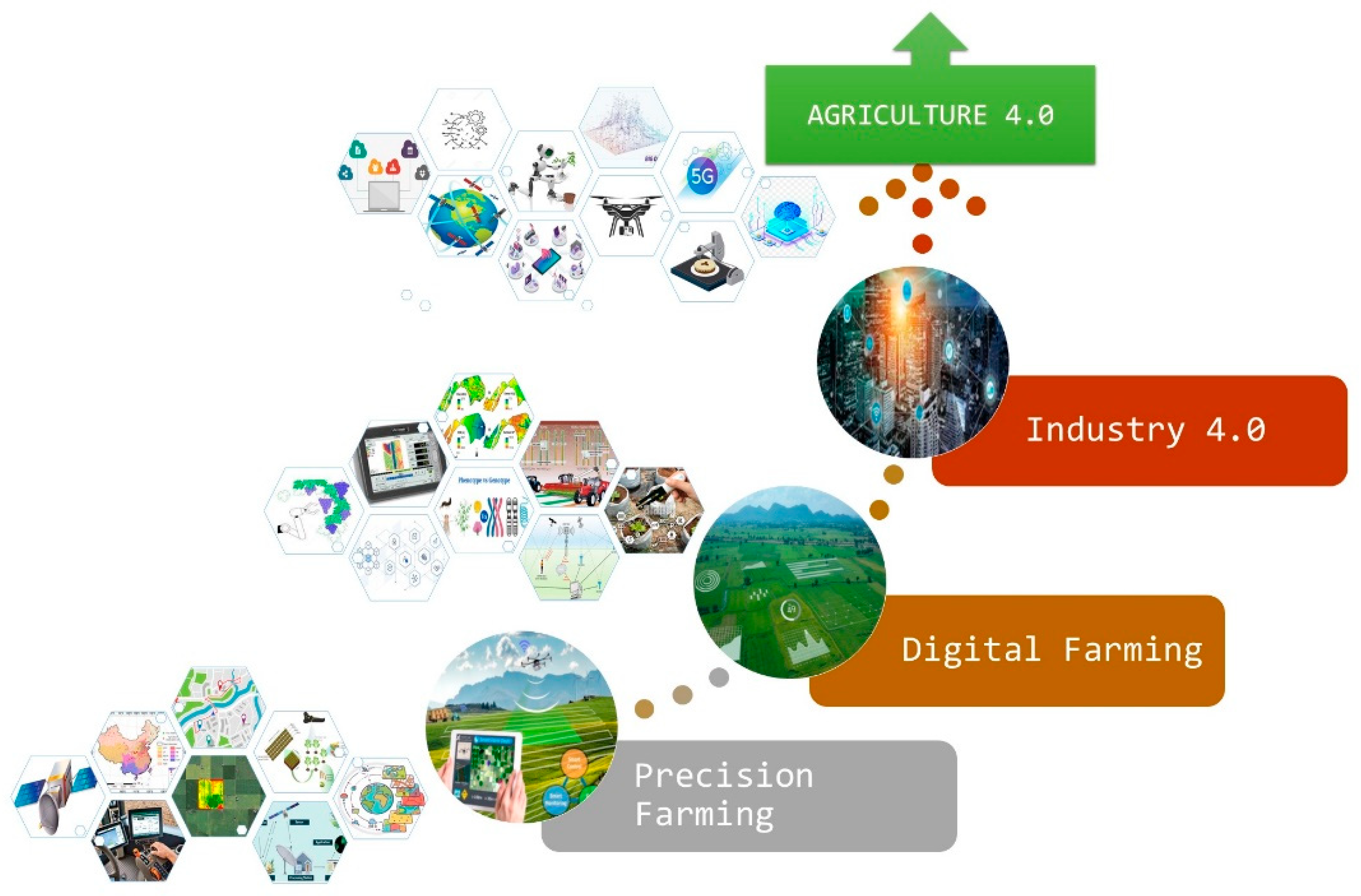}
    \caption{Evolution of trends related to precision agriculture. Source: Karunathilake et al. \cite{karunathilake2023path}.}
    \label{fig:precision_agri_trends}
\end{figure}

The prevision of food needs is a problem that has been studied for centuries as it is a key aspect in the preservation of the human species \cite{malthus1798essay}. Following current trends, the constant growth of the human population requires an increase in food production and precision agriculture can help to achieve this goal by improving the efficiency of farming practices \cite{yahya2018agricultural, corallo2018industry}. The United Nation's world population prospects points out that the population will reach 8.6 billion people by 2030 \cite{un2030prospects}. Precision agriculture consists on the use of technology to optimize crop production and reduce waste. This is crucial in a context where climate change challenges agriculture and food security must be ensured to a larger population \cite{tamasiga2023forecasting}. Accurate localization sustained by reliable sensing sources is crucial for precision agriculture, as it allows farmers to precisely apply fertilizers, pesticides, and other inputs to specific areas of the field \cite{soussi2024smart}. Figure \ref{fig:precision_agri_trends} depicts the evolution on trends related to precision agriculture in a schema proposed by Karunathilake et al. \cite{karunathilake2023path}.

Unmaned aerial vehicles (UAVs) provide high utility in precision agriculture, as they can be used to monitor crop health, identify areas of stress, and optimize irrigation. UAVs are more economic than manned aircrafts, and provide higher quality images with respect to satellites \cite{phang2023satellite, toscano2024unmanned, radoglou2020compilation}. UAV-based remote sensing for precision agriculture has a wide variaty of benefits, such as improved yield prediction, disease detection, and resource management \cite{awais2023uav}. In what regards to real-world applications, in \cite{singh2022intelligent}, the authors propose an Internet-of-Things (IoT) UAV-based system for precision agriculture, focused on solving accurately the path planning problem in order to achieve optimal coverage of the field. In \cite{mukhamediev2023coverage}, a solution to the path planning problem for UAVs in precision agriculture is proposed, using a genetic algorithm approach to optimize the coverage of the field by a heterogeneous group of UAV vehicles.

Precision agriculture benefits from robust DL networks that accurately detect and classify the crops \cite{coulibaly2022deep, kalla2025precision}. These networks can be trained on large datasets of agricultural images, making them able to learn the unique characteristics of different crops and identify them with high accuracy. Reviews such as \cite{saranya2023comparative} analyze the performance of different DL architectures for crop detection and classification, highlighting the strengths and weaknesses of each approach. The importance of IoT applications is also highlighted, as they can provide real-time data on crop health and environmental conditions, allowing farmers to make informed decisions about their farming practices.

\subsubsection{Autonomous agricultural vehicles}\label{subsubsec:autonomous_agri}
\begin{figure}[h]
    \centering
    \includegraphics[width=0.4\textwidth]{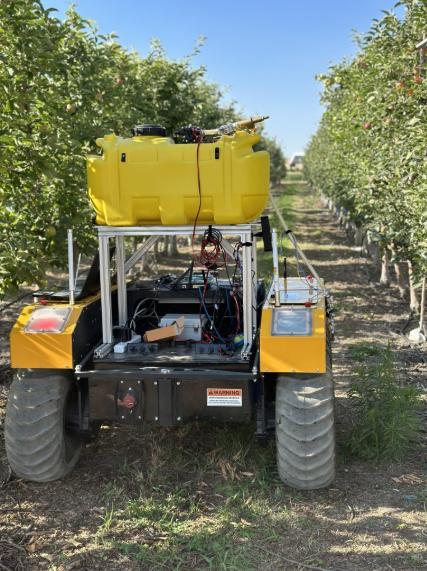}
    \caption{Autonomous agricultural vehicle operating in a orchard environment. Source: Washinton State University \cite{washington2023agricultural}.}
    \label{fig:autonomous_agri_vehicle}
\end{figure}

With a robust solution to the localization problem, several applications can be unlocked in the agricultural field \cite{winterhalter2021localization, martini2022position, ma2025advances}. One of the main applications is on autonomous agricultural vehicles, such as tractors and harvesters \cite{rondelli2022review}. An autonomous agricultural vehicle can perform a variety of tasks, including planting, fertilizing, pruning \cite{silwal2022bumblebee, bhat2025revisiting}, hill farming \cite{padhiary2024enhancing}, and harvesting \cite{xiao2024review, zhou2022intelligent, yoshida2022automated}. In orchard environments, autonomous robots can also be useful in order to transport operators performing tree maintenance, as earlier studies from 2015 \cite{bergerman2015robot}. In order for an autonomous agricultural vehicle to operate efficiently, it needs an optimal sensor setup that allows it to perceive the environment accurately \cite{etezadi2024comprehensive} and a robust solution to the localization problem \cite{hua2025key}. LPR can help these vehicles to recognize previously visited locations and navigate the field more efficiently. Figure \ref{fig:autonomous_agri_vehicle} shows an example of an autonomous agricultural vehicle operating in a vineyard environment.

A reliable path planning solution is crucial for the correct functioning of an autonomous agricultural vehicle \cite{navone2025gps, luo2025navigation}. DeepWay \cite{mazzia2021deepway} is an example of a DL approach that predicts waypoints in vineyard settings for global path generation taking an occupancy grid of the vineyard as an input. By accurately localizing itself within the field, the vehicle can optimize its path and reduce overlap, leading to more efficient operations and reduced fuel consumption \cite{liu2025autonomous}. As an extension to the DeepWay method, Salvetti et al. \cite{salvetti2022waypoint} propose a DL method that clusters waypoints and vineyard rows using a contrastive loss function. The method improves the accuracy of visual SLAM systems in vineyard environments by leveraging the unique structure of rows and waypoints. Through experiments on synthetic data, the authors demonstrate the effectiveness of their approach, showing that their method can accurately classify waypoints. In \cite{aghi2020local}, the authors pioneered a motion planning solution in vineyard environments, leveraging depth maps and convolutional neural networks (CNNs). The method uses depth maps derived from stereo vision to capture the 3D structure of the environment and identify the end of the vineyard rows. In \cite{liu2023vision}, Liu et al. present a solution for the autonomous navigation problem, solving the path planning problem from the data captured by a vision sensor. In order to achieve a robust solution, the authors use segmentation to identify the traversable zones of the environment from the heatmap of an RGB-D image.

\subsubsection{Crop monitoring}\label{subsubsec:crop_monitoring}
Crop monitoring involves the continuous observation of plant growth, health, and productivity over time, often through remote sensing or autonomous field robots \cite{mishra2025smart}. This task is of high importance in order to ensure food security and sustainable agricultural practices, as well as avoiding speculation in food prices \cite{wu2023challenges, karmakar2024crop}. Later trends apply DL techniques to crop monitoring, enabling more accurate and efficient assessments \cite{khaki2019crop, swain2024empowering}. In \cite{omia2023remote}, the authors present a comprehensive review of remote sensing techniques for crop monitoring, discussing the challenges and opportunities in this field. The review covers various aspects of remote sensing for crop monitoring, including data acquisition, processing, and analysis. The authors also discuss the potential benefits of remote sensing for crop monitoring, such as improved yield prediction, disease detection, and resource management \cite{kotwal2023agricultural}. However, while \cite{omia2023remote} provides a high-level overview of remote sensing for monitoring purposes, it does not delve into the algorithmic challenges of spatial awareness. Our work complements this by focusing specifically on LPR, a critical yet often overlooked component for autonomous navigation in complex agricultural rows where traditional GPS or visual data may fail.

While primarily focused on assessing crop status, these activities are closely related to localization. Variations in crop height, density, and spectral appearance throughout the growing season introduce substantial temporal changes that can affect localization performance. For example, navigation systems relying on visual or geometric cues may face difficulties when the vegetation canopy evolves, occluding landmarks or altering the perceived structure of the environment \cite{teng2023multimodal}. In order to provide a solution to the inspection and monitoring of crops using DL techniques methods such as QSeedNet \cite{qiang2025qseednet} or PTL-Inception \cite{gulzar2025ptl} have been proposed. The use of DL techniques has been demonstrated to facilitate more precise and expeditious assessments of crop health and growth. This subject has been extensively examined in recent literature, as evidenced by the numerous reviews conducted on the subject \cite{gadicha2025comparative, sambu2025deep, revathi2025pomegranate}. While these reviews excel at summarizing the SOTA in crop trait analysis, the operational challenge of consistently localizing a platform within those crops remains a separate and complex problem. This study complements the existing literature by providing a specialized review of LPR, a task that differs significantly from health assessment in terms of data geometry, feature extraction, and temporal consistency.

Applying clustering algorithms to these unstructured environments is an effective solution in order to classify distinct regions of interest. In \cite{uyeh2022online}, the authors present an online machine learning-based sensors clustering system for efficient and cost-effective environmental monitoring in controlled environment agriculture. This method uses clustering techniques to group similar sensor readings, enabling more efficient data processing and analysis. In the solution proposed by Swain et al. \cite{swain2024empowering}, the authors propose an unsupervised clustering approach for crop and weed identification in agricultural fields using LiDAR data. The method uses a combination of geometric and spectral features to cluster the point cloud data into different classes, which facilitates accurate identification of crops and weeds.

In \cite{ahmadi2022bonnbot}, BonnBot is presented, an autonomous field robot designed for crop monitoring tasks in agricultural environments. The robot is equipped with a variety of sensors, including LiDAR, cameras, and GPS, to enable accurate localization and navigation within the field. BonnBot is capable of performing a range of crop monitoring tasks, including plant height measurement, leaf area index estimation, and disease detection. The authors demonstrate the effectiveness of BonnBot through field experiments, showing that it can accurately localize itself within the field and perform crop monitoring tasks with high accuracy. 

\subsubsection{Phenotyping}\label{subsubsec:phenotyping}
Phenotyping refers to the process of measuring and analyzing observable traits of plants, such as growth rate, stress, yield, and resistance to pests and diseases \cite{gill2022comprehensive}. This information is crucial for plant breeding programs, as it allows researchers to identify desirable traits and develop new crop varieties that are better suited to specific environments \cite{arya2022deep, atefi2021robotic, akhtar2024unlocking}. Recent sensing technologies have improved the accuracy and efficiency of phenotyping, enabling high-throughput data collection and analysis \cite{wang2025sensors}.

\begin{figure}[h]
    \centering
    \includegraphics[width=0.8\textwidth]{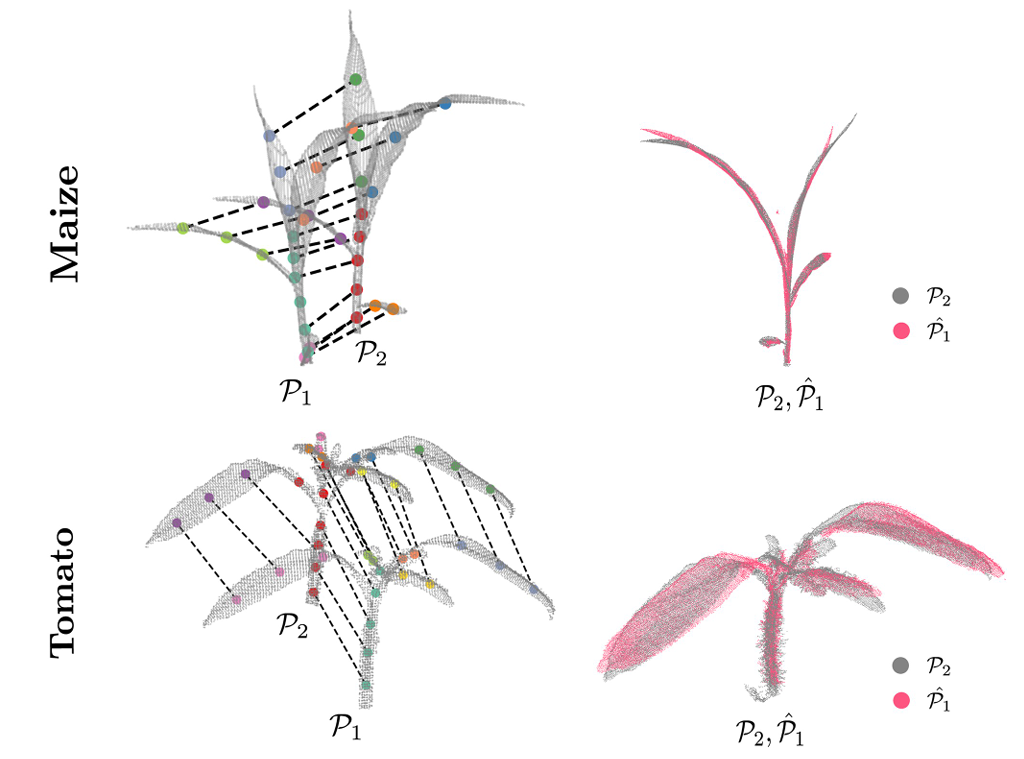}
    \caption{Example of 4D registration in the 3D phenotyping method proposed in \cite{chebrolu2021registration} using LiDAR data from the Pheno4D dataset \cite{schunck2021pheno4d}.}
    \label{fig:phenotyping}
\end{figure}

Multi-temporal registration of LiDAR data is utilized in \cite{chebrolu2021registration} to track plant growth and deformation over several months. By aligning different crop states, this approach enables a precise longitudinal analysis of phenotypic traits. The effectiveness of the method was validated using real-world datasets of maize and tomato, achieving high accuracy in complex registration tasks. The data used for the aforementioned experiments is published under the Pheno4D dataset \cite{schunck2021pheno4d}. Figure \ref{fig:phenotyping} shows an example of the phenotyping process using LiDAR data from the Pheno4D dataset.

To facilitate phenotyping tasks in outdoor environments, Zenkl et al. \cite{zenkl2022outdoor} leverage deep learning—specifically a ResNet-based backbone—to segment plants and extract accurate phenotypic traits from images. A similar goal is shared by the Maize-IAS software \cite{zhou2021maize}, which is tailored for maize plants and automates the extraction of height, leaf area, and biomass through computer vision. While these image-based approaches provide high accuracy for trait analysis, they primarily treat the plant as a static object of study. In contrast, our survey focuses on the geometric understanding of these same environments through LiDAR, shifting the objective from measuring the plant to recognizing the robot's location within the crop rows.

\subsubsection{Pruning and spraying}\label{subsubsec:pruning_spraying}
Pruning and spraying are routine agricultural operations that introduce significant variability in the environment, posing challenges for reliable localization and mapping \cite{gao2006image, silwal2022bumblebee}. Pruning modifies the geometry and appearance of vegetation by removing branches and leaves, which alters canopy density and shape. Such structural changes can disrupt the consistency of spatial features used by localization systems, particularly those relying on visual or geometric cues for map matching \cite{navone2025autonomous}.

\begin{figure}[h]
    \centering
    \includegraphics[width=0.4\textwidth]{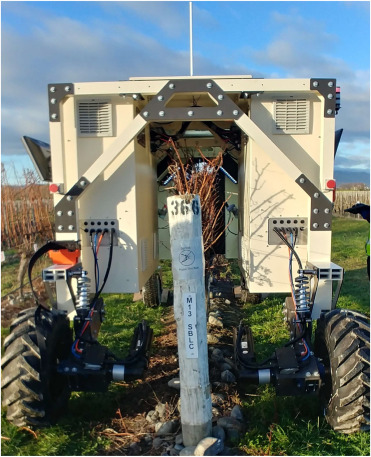}
    \caption{Autonomous pruning robot operating in a vineyard environment proposed by Williams et al. \cite{williams2023modelling}.}
\end{figure}

Spraying operations, on the other hand, affect environmental perception more transiently. The presence of mist, droplets, or wet surfaces can interfere with both optical and LiDAR-based sensing, leading to reflections, occlusions, or altered radiometric responses. Additionally, spraying schedules are often seasonal, introducing recurring but short-lived changes in the sensory data \cite{jiang2024navigation, meshram2022pesticide, lochan2024advancements}.

\begin{figure}[h]
    \centering
    \includegraphics[width=0.8\textwidth]{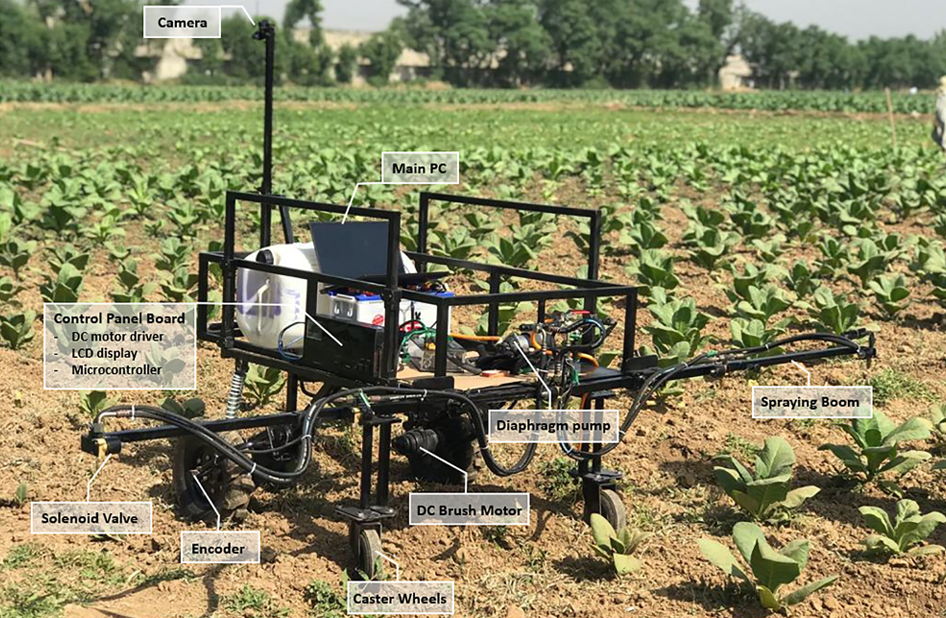}
    \caption{Autonomous spraying robot operating in a tobacco field proposed by Nasir et al. \cite{nasir2023precision}.}
    \label{fig:spraying_robot}
\end{figure}

In \cite{nasir2023precision} a spraying robot for tobacco fields is proposed. Figure \ref{fig:spraying_robot} shows an image of the proposed spraying robot. The robot is equipped with a variety of sensors, including a camera to accurately localize itself within the field and navigate between rows of crops. The robot is capable of performing spraying tasks with high accuracy, reducing the amount of pesticide used and minimizing environmental impact.

Together, pruning and spraying contribute to temporal variability that reduces the reliability of appearance-based localization and map maintenance in agricultural environments. Addressing these challenges requires robust representations capable of handling dynamic vegetation structures and changing surface properties throughout the cultivation cycle.

\section{Localization in agricultural settings}\label{sec:existing_solutions_agri}
Historically, artificial intelligence (AI) research in agricultural environments has predominantly focused on crop monitoring, segmentation, and classification tasks \cite{milioto2018real, moazzam2019review}. This trend is evident in studies ranging from subterranean analysis, such as the multi-stacking ensemble learning for soil classification proposed by Padmapriya et al. \cite{padmapriya2023deep}, to aerial assessments like the UAV-based vineyard segmentation studied by Barros et al. \cite{barros2022multispectral}. While these works are essential for holistic farm management, they primarily rely on spectral or chemical data to describe "what" is in the field. Consequently, the "where"—specifically the problem of spatial awareness and place recognition—has remained comparatively under-explored.

In order to specifically address this type of environments in localization problems, several DL solutions have been released targeting agricultural settings. In this section, we review the existing DL solutions for localization problems (place recognition and SLAM) in agricultural environments. We divide the methods into two main categories: vision-based methods and LiDAR-based methods. Furthermore, we also review methods that incorporate semantic information to improve localization performance. Finally, we comment on the specific challenges that arise in agricultural environments for localization tasks.

\subsection{Vision-based methods}\label{subsec:vision_based_methods}
Focusing on visual descriptors, Tanco et al. \cite{tanco2023learning} developed a self-supervised CNN approach to extract robust features in agricultural terrains. Their method, which includes a modified MobileNet \cite{howard2017mobilenets} for semantic classification, improves visual SLAM performance by adapting keypoint detection to the specific geometry of vineyards. Experimental results from datasets in Northern Italy underscore the effectiveness of this adaptive feature extraction for long-term navigation.

We find a variaty of agricultural SLAM methods as this problem is widely studied as a standard in robotic localization \cite{ding2022recent, khan2025review}. For example, Agri-SLAM \cite{islam2023agri}, a stereo visual SLAM system specifically designed for agricultural environments, concretely tested on a macadamia field. The method introduces a local point and line features recovery technique and combines it with inertial measurements to achieve accurate localization in challenging agricultural settings. In \cite{shu2021slam}, the authors evaluate the performance of monocular visual SLAM in dynamic agricultural environments for the first time, as previous studies focus only on stereo cameras. The study highlights the challenges posed by moving vegetation and changing lighting conditions, and proposes strategies to improve SLAM robustness in such settings. The mask ORB label optimization SLAM (MOLO-SLAM) method \cite{lv2024molo} is another visual simultaneous localization and mapping (VSLAM) system specifically designed for of detecting and eliminating dynamic objects in outdoor forest and tea garden agricultural scenarios. This method combines ORBSLAM2 \cite{mur2017orb} with the Mask-RCNN (region-based convolutional neural network) instance segmentation algorithm to identify and mask dynamic objects in the environment, improving the accuracy of localization and mapping in these challenging settings. Xu et al. \cite{xu2024stereo} propose a robust visual-inertial SLAM algorithm which is based on point-lines features detected on the input stereo images. This visual features are matched to perform a keypoint selection and, subsequently, the pose is stimated using a sliding window algorithm.

\subsection{LiDAR-based methods}\label{subsec:LiDAR_based_methods}
ORCHNet \cite{barros2023orchnet} is a DL application that focuses on LPR in orchards. It proposes a new feature aggregation method for global descriptors that can be applied to different backbones. This proposed method for orchard environments have reported more than 90\% of Recall@1\% on same-season scenarios. However, the Recall@1 value remains lower, with values around 50\%. On a similar note, TriLoc-NetVLAD \cite{sun2024triloc} is another DL approach that enhances long-term place recognition in vineyards using a novel LiDAR-based method. The final approach integrates a handcrafted spatial context descriptor, which is derived from the density, height, and intensity of the point cloud, with a channel selection strategy that prioritizes the most successful layers of the backbone. These layers are more robust to face the unstructured nature of vineyards. This method was tested both against same season and cross season cases, with Recall@1 values of 75\% and 55\%, respectively. Similarly, SPVSoAP3D \cite{barros2024spvsoap3d} is a recent method that introduces a novel 3D point cloud representation called Spherical Projection Voxelization (SPV) for LPR in agricultural environments. The local features of the point cloud are aggregated following a Second-Order Average Pooling (SoAP) strategy. In contrast to this, Barros et al. also proposed PointNetPGAP-SLC \cite{barros2024pointnetpgap}, which is another LiDAR-based DL method that focuses on place recognition in orchards. It was developed by the same authors of SPVSoAP3D, which follows a similar pipeline. PointNet \cite{qi2017pointnet} functions as the foundational framework for the extraction of local features from point clouds. These features are then aggregated through the implementation of a novel Pairwise Feature Interactions and Global Average Pooling (PGAP) algorithm, thereby yielding a global descriptor. The method also incorporates a Segment-Level Consistency model (SLC) to enhance the robustness of the place recognition process. This additional signal, which is taken into account along with the LazyTriplet loss function, considers the row or waypoint each point is located in. This method achieves higher Recall@1 values of 75\%, but was also only tested on similar seasonal campaigns. A significant advancement in orchard-specific navigation is the handcrafted LiDAR-based loop detection method introduced by Ou et al. \cite{ou2023place}. Their approach achieves robust place recognition by encoding point clouds into a Spatial Binary Pattern (SBP), followed by an entropy-based attention reweighting strategy. Although this method demonstrates SOTA performance in large-scale orchards and the KITTI dataset, it represents a specific handcrafted lineage within the field. The most recent LPR DL method MinkUNeXt-VINE \cite{vilellacantos2026lowcosthighefficiency} focuses on achieving robust performance with low-cost, low-resolution inputs. It simplifies the architecture of its parent method, MinkUNeXt \cite{cabrera2025minkunext}, and emphasizes on encoding repetitive environments with lower-dimensional global descriptors. Additionally, it uses a Matryoshka Representation Learning (MRL) \cite{kusupati2022matryoshka} loss function for processing different dimensionalities of the global descriptor and enhance efficiency. This approach achieves nearly 70\% of Recall@1 with a Livox sensor. The metrics reported by these methods tested in agricultural settings are commented in depth in Section \ref{sec:metrics}.

In the context of SLAM methodologies, Aguiar et al. \cite{aguiar2022localization} propose VineSLAM, a LiDAR-based SLAM system that is specifically designed for agricultural environments. The method uses point-feature extraction and semiplanes segmentation to generate a comprehensive map of the environment. Subsequently, the information is integrated into a particle filter, thereby facilitating precise localization. Fei and Vougioukas \cite{fei2022row} present the concept of row templates. These templates are derived from point cloud data, with the associated ground truth values serving as a reference point. These values represent the anticipated measurement values of the sensor (voxel occupancy frequency) in a three-dimensional grid configuration, under the assumption that the sensor is precisely placed at the centerline of the row, with no lateral or heading errors present. Subsequently, the localization process is executed through the use of a conventional Monte Carlo approach. The SG-ISBP framework \cite{ou2024sg} is another LiDAR-based SLAM system that incorporates the density of the point cloud in distinct voxel-defined regions. Specifically, the method involves the integration of the inertial measurement unit (IMU) and a de-skewed version of the point cloud with ground-assisted odometry values to form an improved spatial binary pattern (ISBP) handcrafted loop closing algorithm. This methodology for closing general loops can be applied to distinct localization applications, such as place recognition or SLAM, in isolation. In Purdue-AgSLAM (P-AgSLAM) \cite{kim2024p}, the authors present LiDAR SLAM system specifically designed for robot pose estimation and agricultural monitoring in maize fields. This SLAM approach is designed to address the unique morphological features of maize fields using LiDAR technology, combining a feature extraction mechanism with an Extended Kalman Filter (EKF) fusion of the input orientation sources (wheel odometry and IMU). The P-AgSLAM system is integrated in the Purdue Ag-Bot (P-AgBot) robotic setup, which is equipped with a monocular camera, two 3D LiDAR (Velodyne VLP-16), a GPS system and an IMU. 

\subsection{Semantic information}\label{subsec:semantic_information}
Methods that incorporate semantic information have been suggested as a way to overcome the challenges posed by agricultural environments. One of such methods is Keypoint Semantic Integration (KSI) \cite{de2025keypoint},which integrates semantic information into the keypoint selection process for LPR in outdoor agricultural environments. KSI enriches keypoints with semantic labels, thereby improving the robustness of place recognition algorithms to environmental changes while accounting for the few distinctive features in these environments, such as pipes, poles, and trunks. The presented results demonstrate improved performance on various descriptors and sequences for place recognition and SLAM metrics. In Papadimitriou et al. \cite{papadimitriou2022loop}, the authors propose a method that applies semantic segmentation of grape vines. This method uses a Mask-RCNN DL approach and uses the trunks as a reference for localizing vineyards. The method's output is combined with a traditional Graph-SLAM localization algorithm to improve localization accuracy.

Other approaches such as \cite{de2025particle} have proposed novel particle filter-based localization methods that incorporate semantic information to enhance the robustness of the methods in agricultural environments. By leveraging semantic information, these methods can more effectively address the challenges posed by unstructured and dynamic agricultural settings. This method uses semantic information to enhance the particle filter likelihood estimation process. The proposed approach integrates semantic labels into the localization framework, enhancing the robustness and accuracy of SLAM in agricultural environments where traditional methods may encounter challenges due to the absence of distinctive features and dynamic changes in the environment.


As a summary, Table \ref{tab:agri_lpr_methods} presents the commented localization methods designed within an agricultural context. This table includes the method name, type of sensor took as an input, year of publication, the approach followed (wether it is a place recognition or a SLAM method), and key features of each approach.

\begin{landscape}
    \begin{table}[h]
        \centering
        \caption{Summary of the main localization methods specifically designed for agricultural environments.}
        \begin{tabular}{|c|c|c|c|c|}
            \toprule
            \textbf{Type} & \textbf{Method} & \textbf{Year} & \textbf{Approach} & \textbf{Base architecture} \\
            \midrule
            \multirow{6}{*}{Vision} & Shu et al. \cite{shu2021slam} & 2021 & SLAM & ORB-SLAM2 \cite{mur2017orb} \\
            & Papadimitriou et al. \cite{papadimitriou2022loop} & 2022 & SLAM & Mask R-CNN + Graph-SLAM \\
            & Agri-SLAM \cite{islam2023agri} & 2023 & SLAM & Pose graph optimization (PGO) \\
            & Xu et al. \cite{xu2024stereo} & 2024 & SLAM & Sliding window \\
            & Lv et al. \cite{lv2024molo} & 2024 & SLAM & Mask-RCNN \\
            & KSI \cite{de2025keypoint} & 2025 & Localization & YOLO + SuperGlue (matching) \\
            \hline
            \multirow{10}{*}{LiDAR} & VineSLAM \cite{aguiar2022localization} & 2022 & SLAM & Particle filter \\
            & Fei and Vougioukas \cite{fei2022row} & 2022 & SLAM & Monte Carlo \\ 
            & ORCHNet \cite{barros2023orchnet} & 2023 & PR & Global feature aggregation \\
            & Ou et al. \cite{ou2023place} & 2023 & PR & Attention score map + Spatial Binary Pattern \\
            & TriLoc-NetVLAD \cite{sun2024triloc} & 2024 & PR & CNN + NetVLAD \\
            & SPVSoAP3D \cite{barros2024spvsoap3d} & 2024 & PR & Spherical Projection Voxelization representation and Second Order Average Pooling \\
            & PointNetPGAP-SLC \cite{barros2024pointnetpgap} & 2024 & PR & Pairwise Feature Interactions and Segment-Level Consistency \\
            & SG-ISBP \cite{ou2024sg} & 2024 & PR \& SLAM & Improved Spatial Binary Pattern \\
            & P-AgSLAM \cite{kim2024p} & 2024 & SLAM & Feature Matching + Extended Kalman Filter \\
            & De et al. \cite{de2025particle} & 2025 & SLAM & Semantics + Particle Filter Likelihood Estimation \\
            \bottomrule
        \end{tabular}
        \label{tab:agri_lpr_methods}
    \end{table}
\end{landscape}


\subsection{Specific challenges}\label{subsec:challenges_agri}
While urban environments are characterized by distinct and stable features such as buildings and roads, agricultural environments present unique challenges due to their unstructured and dynamic nature. Table \ref{tab:urban_vs_agri} summarizes the main differences between urban and agricultural environments. These differences pose significant challenges for LPR algorithms, which must be able to adapt to the specific characteristics of agricultural settings.

\begin{table}[h]
    \centering
    \caption{Comparison between urban and agricultural environments for LPR.}
    \begin{tabular}{|c|c|c|}
        \toprule
        \textbf{Aspect} & \textbf{Urban Environments} & \textbf{Agricultural Environments} \\
        \midrule
        Distinctive features & Presence of buildings, roads & \makecell{Lack of distinctive features; \\ similar-looking plants} \\
        \midrule
        Repetitive patterns & Less common; unique structures & \makecell{Common; rows of crops or trees} \\
        \midrule
        Seasonal changes & Less pronounced; stable structures & \makecell{Significant changes; \\ vegetation growth and decay} \\
        \midrule
        Environmental variability & More structured and predictable & \makecell{Unstructured and dynamic} \\
        \bottomrule
    \end{tabular}
    \label{tab:urban_vs_agri}
\end{table}

\subsubsection{Lack of distinctive features}\label{subsubsec:lack_features}
Agricultural environments often lack distinctive features that can be easily recognized by place recognition algorithms. For example, in a field of crops, the plants may look similar and lack unique characteristics that can be used for recognition. This can lead to confusion and misidentification of locations.

\begin{figure}[h]
    \centering
    \includegraphics[width=0.8\textwidth]{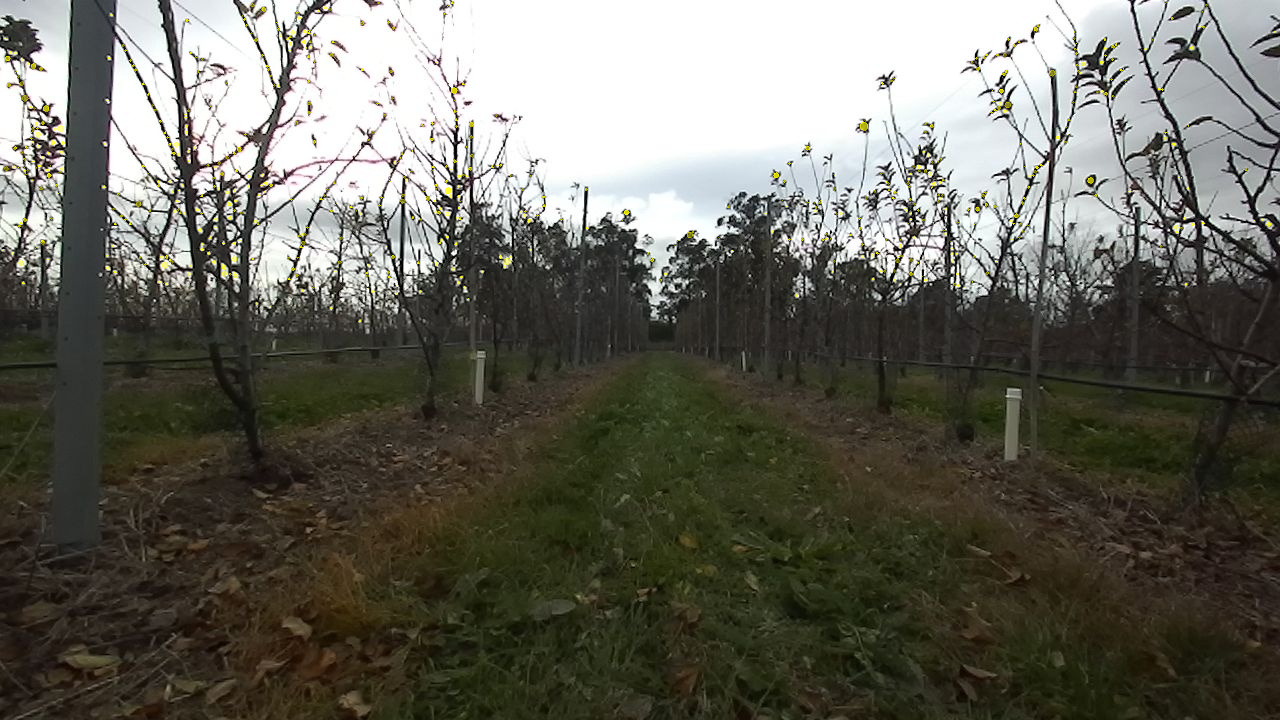}
    \caption{Example of lack of distinctive features in an orchard environment. The plants look identic and lack unique characteristics that can be used for recognition. Image taken from the MAgro dataset \cite{marzoa2024magro}.}
    \label{fig:lack_features}
\end{figure}

\subsubsection{Repetitive patterns}\label{subsubsec:repetitive_patterns}
Agricultural environments often contain repetitive patterns, such as rows of crops or trees, which can make it difficult for place recognition algorithms to distinguish between different locations. This can lead to false positives, where the algorithm incorrectly identifies a location as being the same as another location.

\begin{figure}[h]
    \centering
    \includegraphics[width=0.8\textwidth]{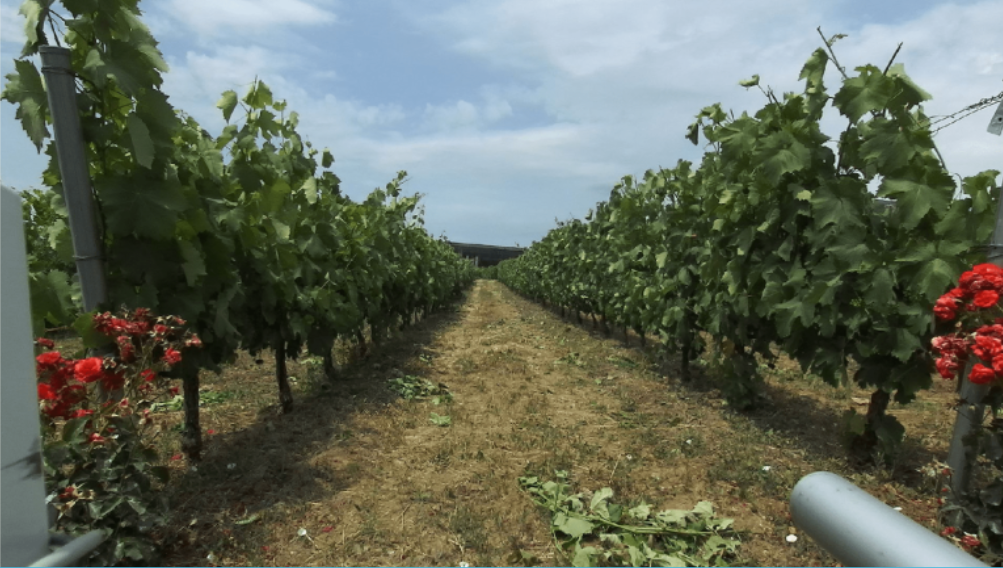}
    \caption{Example of repetitive patterns in a vineyard environment. The rows create a repetitive pattern that can be challenging for place recognition algorithms to accurately identify locations. Image obtained from the BLT dataset \cite{polvara2024bacchus}.}
    \label{fig:repetitive_patterns}
\end{figure}

Figure \ref{fig:repetitive_patterns} shows an example of repetitive patterns in an orchard environment. The rows of trees create a repetitive pattern that can be challenging for place recognition algorithms to accurately identify locations.

\subsubsection{Seasonal changes}\label{subsubsec:seasonal_changes}
\begin{figure}[h]
    \centering
    \begin{subfigure}{0.45\textwidth}
        \centering
        \includegraphics[width=\textwidth]{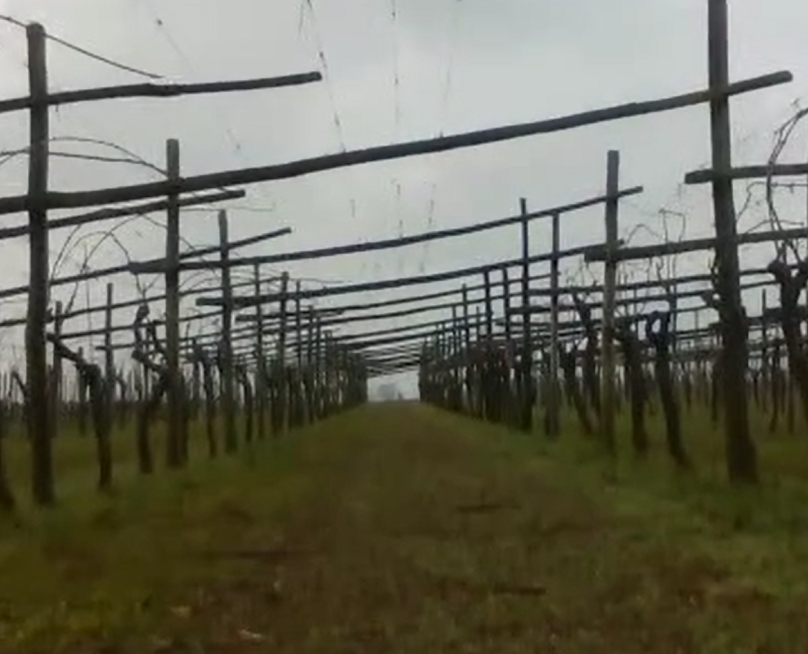}
        \caption{Winter season}
    \end{subfigure}
    \hfill
    \begin{subfigure}{0.45\textwidth}
        \centering
        \includegraphics[width=\textwidth]{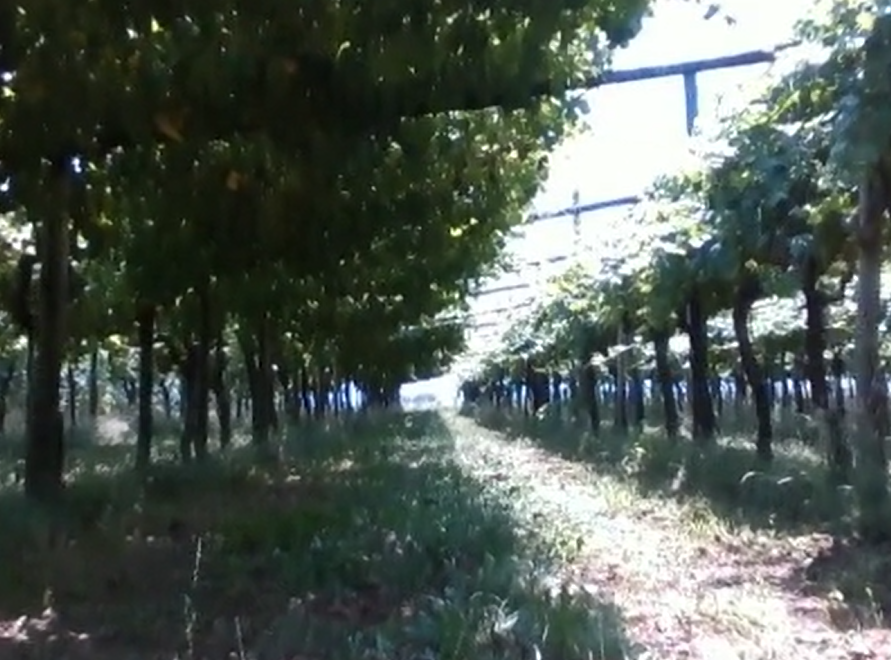}
        \caption{Summer season}
    \end{subfigure}
    \caption{Example of seasonal changes from a same location in a pergola-type vineyard environment extracted from the TEMPO-VINE dataset. (a)  Winter season with bare vines. (b)  Summer season with abundant vegetation covering the robot.}
    \label{fig:seasonal_changes}
\end{figure}

Agricultural environments are subject to seasonal changes, which can significantly alter the appearance of the environment. For example, a field of crops may look very different in the summer compared to the winter, when it loses all of the leaves. This can make it challenging for place recognition algorithms to accurately identify locations, as the features used for recognition may no longer be present. Even to an external observator, the place may look completely different depending on the season. Figure \ref{fig:seasonal_changes} shows an example of seasonal changes in a vineyard environment (the TEMPO-VINE dataset). It illustrates how the same location can appear drastically different between winter and summer seasons, posing a significant challenge for LPR systems.

\subsubsection{Irregular terrain}\label{subsubsec:irregular_terrain}
In contrast to urban environments, agricultural settings are characterized by irregular terrain, a factor that can introduce navigational challenges for robotic systems. The presence of variations in elevation, uneven ground surfaces, and the existence of natural obstacles, such as rocks or ditches, can collectively contribute to a more complex scenario for the implementation of path planning applications. The odometry of robots operating in such environments, where the terrain may be slippery or unstable, can be adversely affected, leading to inaccuracies in position estimation. These irregularities can result in deviations from the intended path, thereby complicating the localization process. 

\subsubsection{Geo-referencing difficulties}\label{subsubsec:georeferencing_difficulties}
Agricultural environments frequently present challenges for geo-referencing due to the absence of reliable GPS signals. The obstruction of satellite signals by factors such as dense vegetation or the presence of large structures, such as barns or silos, can be attributed to multipath effects. Consequently, obtaining a reliable ground truth for localization can be challenging in such environments.

\section{The LiDAR Place Recognition problem}\label{sec:LPR_problem}
LPR constitutes a crucial component of long-term localization and mapping systems in autonomous robotics. The objective is to determine whether a 3D LiDAR scan corresponds to a previously visited location, enabling loop closure detection and global position correction. Compared to vision-based approaches, LiDAR offers high geometric accuracy and invariance to lighting and appearance changes, which is essential for operation under varying environmental conditions. Nevertheless, the unstructured nature of point clouds and the presence of dynamic or seasonal changes make the extraction of robust and distinctive representations a challenging task. Recent advances in DL have significantly improved LiDAR-based place recognition by learning compact and invariant descriptors capable of generalizing across different scenes and sensor configurations \cite{luo2023bevplace, kong2024sc_lpr, xia2023casspr}.

\begin{figure}[h]
    \centering
    \includegraphics[width=0.8\textwidth]{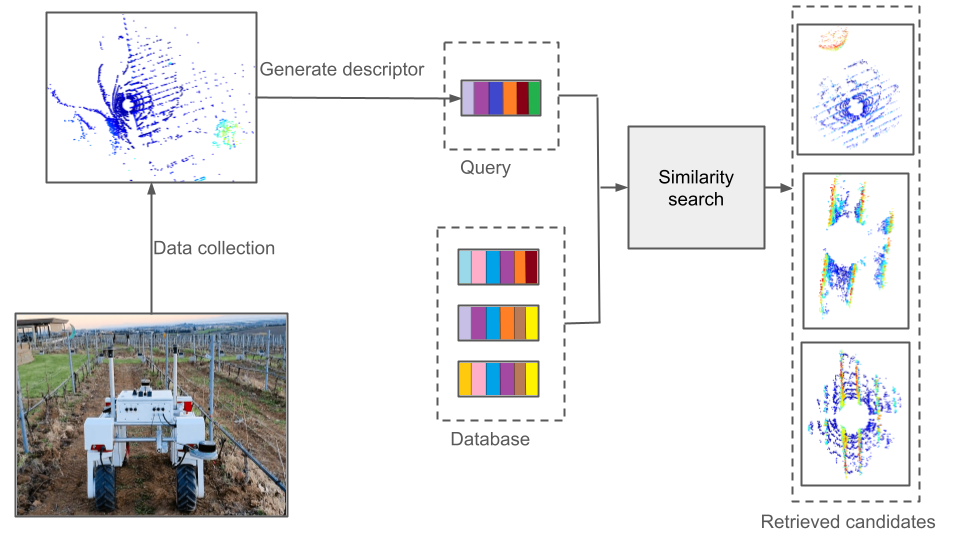}
    \caption{Diagram of a typical LPR pipeline. For each point cloud, a global descriptor is generated. This descriptor can then be used for place recognition by comparing it to a database of previously stored descriptors.}
    \label{fig:DL_LPR_diagram}
\end{figure}

Figure \ref{fig:DL_LPR_diagram} shows a typical LPR pipeline. The input point cloud can be first preprocessed to remove noise and outliers or in order to select a region of interest, and then a feature extraction network is used to extract local features from the point cloud. These local features are then aggregated into a global descriptor using a feature aggregation method. Finally, the global descriptor is compared to a database of previously stored descriptors to recognize the place. This resulting descriptor can then be used for localization within a pre-built map of the environment or for other applications, such as clustering of the environment.

\subsection{The place recognition problem}\label{subsec:PR_problem}
Neuroscience has studied the ability of animals to recognize places for decades. Early studies from the decade of 1940 introduce the concept of cognitive map. A cognitive map is a mental representation of the environment that permits animals to navigate and recognize places. Studies such as those from Tolman in 1948 \cite{tolman1948cognitive} laid the groundwork for understanding how animals navigate their environment using cognitive maps. In this study, Tolman demonstrated that rats could learn to navigate a maze by forming a mental representation of the environment, rather than simply memorizing a sequence of movements. This finding challenged the prevailing behaviorist view of learning and paved the way for further research into the neural mechanisms underlying spatial cognition. 

While earlier studies discovered and analyzed cognitive maps in animal behavior, it was not until the 1970s that the neural basis of this ability was discovered. The discovery of place cells in the hippocampus of rats by O'Keefe and Dostrovsky in 1971 \cite{o1971hippocampus} represented a significant breakthrough in understanding how animals navigate their environment. Place cells are neurons that become active when an animal is in a specific location, suggesting that they play a crucial role in spatial memory and navigation. These findings have been extended to other species, including humans, and they have also led to a deeper understanding of the neural mechanisms underlying spatial cognition. The study of place cells has also inspired the development of computational models for place recognition in robotics and AI. 

On the human behaviour side, studies starting from the decade of 1970 analyzed how humans form mental representations of their environment and use them to navigate \cite{graham1976mental}. Other studies have been adding information to the place recognition problem from the human neuroscientific and geographical behavioural perspectives in the following decades \cite{golledge1992place}. All of these studies showed that humans rely on a variety of cues, including landmarks, spatial relationships, and environmental features, to create cognitive maps of their surroundings.

\subsection{LiDAR Deep Learning methods}\label{subsec:deeplearning_rw}
The objective of using trainable methods to find a solution problem is to learn the embeddings of the input data, in this case, point clouds, into a lower-dimensional space where similar places are closer together. This is typically achieved using deep neural networks, which can learn complex representations of the input data. Deep learning methods can be supervised or unsupervised. Supervised methods require a labeled dataset, where each point cloud is associated with a specific location. The network is trained to minimize the distance between the embeddings of point clouds from the same location and maximize the distance between embeddings from different locations. Unsupervised methods, on the other hand, do not require labeled data and instead they rely on techniques such as clustering or autoencoders to learn the embeddings.

PointNetVLAD \cite{uy2018pointnetvlad} is one of the most popular DL methods for LPR. It combines the PointNet architecture \cite{qi2017pointnet} for feature extraction with the NetVLAD layer \cite{arandjelovic2016netvlad} for feature aggregation. The PointNet architecture is designed to handle unordered point clouds and can it learn local features from the input data. The NetVLAD layer aggregates these local features into a global descriptor that can be used for place recognition. The network is trained using a triplet loss function, which encourages the embeddings of point clouds from the same location to be closer together than those from different locations. This primary approach offers 80\% of Recall@1\% in the Oxford Robotcar Dataset \cite{maddern20171}. Other popular methods include LPD-Net \cite{liu2019lpd}, which builds upon PointNetVLAD by incorporating local features and a graph-based feature aggregation method. OverlapNet \cite{chen2022overlapnet} is a method that uses a Siamese network architecture to learn the embeddings. The network is trained to predict the overlap between two point clouds, which can be used for place recognition. MinkLoc3D \cite{komorowski2021minkloc3d} is another method that uses sparse 3D convolutions to learn the embeddings. This method is designed to handle large-scale point clouds and can achieve high accuracy with low computational cost. The authors also proposed MinkLoc3Dv2 \cite{komorowski2022improving}, which uses a ranking-based loss function instead of the previously used triplet one along with structural modifications. These changes improves the recall results with respect to the previous version of the framework by 3.4\%. On a similar note, MinkUNeXt \cite{cabrera2025minkunext} is a recent method that builds upon MinkLoc3D by incorporating a UNet-like architecture for feature extraction and a ranking-based loss function. This method has shown promising results in large-scale outdoor environments, with Recall@1 results above 90\% in several urban datasets.

Recent approaches have also studied the use of additional information, such as intensity or color, to improve the performance of LPR. For example, the variants MinkLoc3D-SI \cite{zywanowski2021minkloc3d} and MinkUNeXt-SI \cite{vilella2025minkunext} have also proved to present superior results to their original backbones by using the spherical projection of the points and including the intensity information, with recall improvements of around 5\%. Cgis-Net \cite{ming2022cgis} is another method that incorporates color information into the point cloud data. This method adds color and geometric features to semantic features, which are then used to learn the embeddings. Aiming at including semantic information, SegMap \cite{dube2020segmap} is a method that uses a 3D convolutional neural network to learn the embeddings of segmented point clouds. The network is trained to predict the semantic labels of the segments, which can be used for place recognition. Similarly, Coral \cite{pan2021coral} uses the bird-eye view (BEV) image fused with vision information that semantically enriches the output descriptor. Locus \cite{vidanapathirana2021locus} is another method that incorporates semantic information into the point cloud data. This method uses aggregated high-level semantic features with low-level geometric features to learn the embeddings, producing a robust global descriptor.

Novel approaches consider incremental learning techniques to adapt the model to new environments or conditions without forgetting previously learned information. For example, InCloud \cite{knights2022incloud} is an incremental learning method that uses a memory buffer to store previously seen data. The model is updated with new data while retaining the knowledge from the memory buffer.

\subsection{LiDAR handcrafted methods}\label{subsec:handcrafted_rw}
Handcrafted methods for LPR typically involve the extraction of geometric features from point clouds, which are then used to create descriptors that can be compared to recognize places \cite{tombari2011combined}. There are different types of handcrafted descriptors, but the most prominent ones are signature-based and histogram-based methods. Firstly, signature-based methods are characterized for encoding the prominent features or repetitive structures of the pointcloud in a signature. Specifically, signature-based methods compress the spatial information of a point cloud into distinctive descriptors, often taking the form of range images, polar projections, or BEV representations. By doing so, they transform unstructured 3D data into a structured format that facilitates efficient matching. The most popular signature-based approach is Scan Context \cite{kim2018scan}, which creates a global descriptor by projecting the point cloud onto a 2D polar grid and encoding the height information. Several variants of Scan Context has been released. For example, Intensity Scan Context \cite{wang2020intensity} incorporates intensity information into the Scan Context descriptor in order to improve its robustness. Scan Context++ \cite{kim2021scan}, on the other hand, improves the robustness of the descriptor to rotation and lateral variations. The results of applying Scan Context to agricultural datasets are presented in the TEMPO-VINE dataset work \cite{martini2025tempo}. Martini et al. demonstrated that, although promising performance can be achieved, performance significantly decreases when seasonal changes are considered, dropping recall values from 90\% to 40\%.

On the other hand, histogram-based methods are those who encode point clouds according to a specific characteristic. They characterise distributions of local or global measurements (distances, normals, curvatures, intensities) as histograms or aggregated statistics. Instances of this type of methods include the Fast Point Feature Histogram (FPFH) \cite{rusu2009fast}, which computes histograms of local geometric features around each point in the point cloud. The histograms are then aggregated into a global descriptor that can be used for place recognition. Another example is the Ensemble of Shape Functions (ESF) \cite{wohlkinger2011ensemble}, which computes a set of shape functions that capture the global geometry of the point cloud. These shape functions are then integrated into a global descriptor that can be used for place recognition. In Omni Point \cite{im2024omni}, the authors propose a novel histogram-based descriptor that captures the omni-directional geometry of the point cloud. The method computes histograms of local geometric features in multiple directions and aggregates them into a global descriptor that can be used for place recognition.


\subsection{Point cloud preprocessing}\label{subsec:preprocessing_rw}
Data preprocessing is a crucial step in LPR, as it can significantly impact the performance of the algorithms. Preprocessing techniques can include filtering, downsampling, and normalization of the point cloud data. 

Filtering techniques can be used to remove noise and outliers from the point cloud, which can improve the accuracy of the place recognition algorithms. Another common filtering approach is to remove the floor points or the points outside a given radius, in order to take only relevant points into account \cite{santo2025ground}. Downsampling techniques can be used to reduce the size of the point cloud, which can improve the computational efficiency of the algorithms and also reduce the memory footprint. Normalization techniques can be used to standardize the point cloud data, which can improve the robustness of the algorithms to changes in scale and orientation.

Another preprocessing technique that has been proved succesful in LPR is the use of other information presented in the LiDAR data, mainly the intensity value of each coordinate. The intensity value represents the intensity of the reflected laser beam and can provide additional information about the surface properties of the scanned objects. Several works have shown that including this information in the input data can improve the performance of LPR algorithms \cite{zywanowski2021minkloc3d, vilella2025minkunext}.

\section{Datasets}\label{sec:datasets}
A variety of datasets have been used to evaluate the performance of LPR algorithms. However, most of these datasets are collected in urban environments, which may not be representative of the challenges faced in agricultural settings. In this section, we analyze some of the most commonly used datasets for LPR and discuss their relevance to agricultural environments. 

\subsection{Transitional datasets}\label{subsec:transitional_datasets}
Although many localization datasets are collected in urban environment, some of these environments include vegeation-rich sections. The KITTI dataset \cite{geiger2013vision} is one of the most widely used datasets for LPR. It includes a variety of urban environments, including some sections with vegetation, such as parks and tree-lined streets. However, the majority of the dataset is collected in urban settings, which may not be representative of the challenges faced in agricultural environments. While a variaty of different trajectories has been captured within this dataset, it does not contain long-term information, as all the sessions were collected within a scope of 1 month. Another popular dataset is the Oxford Robotcar Dataset \cite{maddern20171}, which includes a variety of urban environments, including some sections with vegetation. The total extension of the data captured by this dataset is of 10 km. Although it contains long-term information as the sessions were recorded over a year, the vegetation sections are limited and may not be representative of agricultural settings.

Several datasets capture off-road environments with the presence of natural elements. The ORFD dataset \cite{min2022orfd} is a dataset collected in a mixed urban and off-road (forest) environment. It includes a variety of vegetation types, such as trees, bushes, and grass. The dataset contains long-term information, as the sessions were recorded covering seasons from spring to winter. However, the extension of the recorded data in each session is very limited and the off-road sections are limited and may not be representative of agricultural settings. The Wild-Places dataset \cite{knights2023wild} is a dataset proposed for LPR and SLAM specifically, and was collected in a variety of outdoor environments, including forests, parks, and rural areas. The dataset contains long-term information, as the sessions were recorded over a year. However, the dataset does not include agricultural environments specifically. Finally, the Rellis-3D dataset \cite{jiang2021rellis} is a dataset collected in rural environments, including fields and forests. The dataset contains a variety of vegetation types, including crops and trees. However, the dataset does not include long-term information, as all the sessions were collected within a short period of time. In these datasets, both 3D information from terrestrial LiDAR sensors and accurate ground truth annotations are provided, which are essential for developing and evaluating LPR algorithms.

\subsection{Datasets for agricultural environments}\label{subsec:agri_datasets}
Several datasets have been collected in agricultural environments to address the unique challenges posed by these settings. Table \ref{tab:datasets} summarizes the main characteristics of these datasets, including the year of publication, the crop type, the sensor used, the number of sequences, and whether they include long-term variations.

\begin{landscape}
\begin{table}[htbp]
    \centering
    \scriptsize
    \setlength{\tabcolsep}{3pt}
    \renewcommand{\arraystretch}{0.25}
    \caption{Summary of available agricultural environment datasets.}
    \begin{tabularx}{\linewidth}{c|c|L|L|L|L|c|c|c}
        \toprule
        \textbf{Task} & \textbf{Year} & \textbf{Dataset} & \textbf{Crop type} & \textbf{LiDAR} & \textbf{Vision} & \textbf{GPS} & \textbf{Sequences} & \textbf{Long-term}\\ 
        \midrule
        \multirow{7}{*}{\makecell{Detect. \\ Segment. \\ Count.}} 
        & 2016 & Stein et al. \cite{stein2016image} & Mango & Velodyne HDL64E & \makecell[tl]{Prosilica GT3300c \\ Kowa LM8CX} & Yes & 3 & No\\
        \addlinespace
        & 2017 & Bargoti et al. \cite{bargoti2017deep} & \makecell{Apple \\ Mango \\ Almond} & - & \makecell[tl]{PointGrey LadyBug \\ Prosilica GT3300c \\ Canon EOS60D} & No & 3 & No\\
        \addlinespace
        & 2018 & Liu et al. \cite{liu2018robust} & \makecell{Apple\\Orange} & - & Samsung Galaxy S4 & No & 2 & No\\
        \addlinespace
        & 2019 & CropDeep \cite{zheng2019cropdeep} & Diverse (greenhouse) & - & \makecell[tl]{IoT camera \\ Autonomous robots \\ Smartphone} & No & Unspecified & No\\
        \addlinespace
        & 2020 & MinneApple \cite{hani2020minneapple} & Apple & - & Samsung Galaxy S4 & No & Unspecified & Yes\\
        \addlinespace
        & 2021 & Alessandrini et al. \cite{alessandrini2021grapevine} & Grapevine & - & \makecell[tl]{iPadPro \\ Samsung J7 \\ iPhone 8} & No & Unspecified & No\\
        \addlinespace
        & 2021 & Abdelghafour et al. \cite{abdelghafour2021annotated} & Grapevine & - & Basler Ace (acA2500-14gc GigE) & No & Unspecified & No\\
        \addlinespace
        & 2023 & Low-Light \cite{islam2023agri} & Macadamia & - & Zed Stereo & No & 1 & No\\
        \addlinespace
        & 2023 & GrapeNet \cite{barbole2023grapesnet} & Grapevine & - & \makecell[tl]{One Plus 7 Mobile \\ Intel Real-sense \\D435I } & No & 18 & Yes\\
        \addlinespace
        & 2025 & Mohammed et al. \cite{mohammed2025pomegranate} & Pomegranate & - & Sony ILCE-7RM4 & No & 1 & No\\
        \midrule
        \multirow{8}{*}{Localiz.} 
        & 2016 & Sugarbeets \cite{chebrolu2017agricultural} & Sugarbeet & Velodyne VLP16 Puck & \makecell[tl]{JAI AD-130GE \\ Kinect One} & Yes & 30 & No\\
        \addlinespace
        & 2022 & GREENBOT \cite{canadas2024multimodal} & Tomato & Velodyne VLP-16 & Bumblebee BB2-08S2 & Yes & 9 & No\\
        \addlinespace
        & 2022-2023 & MAgro \cite{marzoa2024magro} & Apple; Pear & Velodyne Puck 3D & StereoLabs Zed2 (2) & Yes & 9 & Yes\\
        \addlinespace
        & 2022-2023 & HORTO-3DLM \cite{barros2024spvsoap3d} & \makecell[tl]{Apple; Strawberry; \\ Cherry; Tomato} & Velodyne VLP-16 & - & Yes & 6 & No\\
        \addlinespace
        & 2022-2023 & BLT \cite{polvara2024bacchus} & Grapevine & Ouster OS1-16 & StereoLabs Zed2 & Yes & 15 & Yes\\
        \addlinespace
        & 2023 & ROSARIO \cite{soncini2025rosario} & Soybean & Velodyne VLP-16 & Intel Realsense D435i & Yes & 6 & Yes\\
        \addlinespace
        & 2023 & ARD-VO \cite{crocetti2023ard} & Vineyards & Velodyne VLP-16 & \makecell[tl]{RedEdge MX \\ Blackfly S} & Yes & 11 & Yes\\
        \addlinespace
        & 2025 & TEMPO-VINE \cite{martini2025tempo} & Grapevine & \makecell[tl]{Velodyne VLP-16 \\ Livox Mid360} & RGB-D Intel Realsense D435 & Yes & 11 & Yes\\
        \bottomrule
    \end{tabularx}
    \label{tab:datasets}
\end{table}
\end{landscape}

The majority of agricultural datasets focus on collecting visual information with RGB cameras. However, there has been a growing interest in collecting LiDAR data in agricultural settings in recent years. The GREENBOT dataset \cite{canadas2024multimodal} is one of the first datasets to include LiDAR data collected in a tomato greenhouse environment. The MAgro dataset \cite{marzoa2024magro} includes LiDAR data collected in apple and pear orchards, while the HORTO-3DLM dataset \cite{barros2024spvsoap3d} includes data from multiple crop types, including apple, strawberry, cherry, and tomato and it was collected using a different sensor setup in each sequence and location. The BLT dataset \cite{polvara2024bacchus} focuses on grapevine environments, while the ROSARIO dataset \cite{soncini2025rosario} includes soybean fields. The ARD-VO dataset \cite{crocetti2023ard} also focuses on vineyards. Finally, the TEMPO-VINE dataset \cite{martini2025tempo} includes long-term LiDAR data collected in vineyards using both Velodyne VLP-16 and Livox Mid360 sensors, providing heterogeneous LiDAR information.

If we compare these datasets with the most popular urban datasets, such as the Ford Campus dataset \cite{pandey2011ford} or the University of Michigan's North Campus Long-Term dataset (NCLT) \cite{carlevaris2016university}, we can notice that the agricultural datasets are still limited in terms of the number of sequences and the variety of environments. There is a wider niche of urban datasets that ease the task of training and evaluating LPR algorithms. However, the unique challenges posed by agricultural environments require specialized datasets that can effectively capture the unstructured nature of these settings. 

It is also important to take into account that the presence of the necessary information for developing LPR applications is a more recent approach, as there are many agricultural environments that have not yet been taken for this precise purpose. The most popular approach is to publish data collected from vision sensors, which facilitates classification algorithms, useful for different tasks such as disease identification \cite{gatou2024artificial} or fruit detection \cite{sa2016deepfruits}. This trend has been explored since the mid 2010s decade, with published data such as \cite{bargoti2017deep}, where the authors propose data taken from different vision systems in different orchards (apple, mango and almond planctations) and they use this data in their proposed Faster RCNN DL detection method. On a similar manner, Liu et al. \cite{liu2018robust} propose an oranges and apples monocular images dataset for fruit counting through their pipeline, and Stein et al \cite{stein2016image} publish data taken in a mango orchard for fruit detection. CropDeep \cite{zheng2019cropdeep} is an annotated dataset created specifically for classification and detection in agricultural environments collected both with mobile cameras, autonomous robots and IoT cameras.

More recent works such as the Low-Light dataset \cite{islam2023agri} presents data collected in macadamia fields, but it does not contain neither LiDAR nor GPS information. The GrapeNet dataset \cite{barbole2023grapesnet} provides RGB and RGB-D information in a vineyard environment in order to support vision applications. The MinneApple dataset \cite{hani2020minneapple} provides annotated visual information taken in an apple orchard, which also supports DL semantic and classification algorithms. The use of UAV LiDAR has also been an extended practice when it comes to collect data in agricultural fields \cite{bouguettaya2022deep}. The VineLiDAR dataset \cite{velez2023vinelidar} provides high-resolution UAV-LiDAR data collected over two years in northern Spain. Alessandrini et al. \cite{alessandrini2021grapevine} propose a vineyards dataset with RGB-D data captured with smartphone devices for detection of the esca disease in leaves with DL applications. Abdelghafour et al. \cite{abdelghafour2021annotated} also provide an annotated imagery dataset taken in a vineyard environment that facilitates the task of identifying diseases in the Merlot grape variaty. In their more recent work, Mohammed et al. \cite{mohammed2025pomegranate} presented a public dataset comprising pomegranate images. These images, which depict both healthy and unhealthy crops, are designed to support the development of advanced learning applications focused on disease detection. Further efforts are needed to collect and curate datasets that can support the development of robust LPR algorithms for agricultural applications, providing data from terrestial LiDAR sensors and accurate ground truth annotations.

\section{Evaluation Metrics}\label{sec:metrics}
The performance of LPR algorithms is evaluated by measuring the ability of the algorithm to correctly recognize previously visited places and to accurately localize the robot within a pre-built map of the environment. In this section, an examination of the most commonly used metrics for evaluating the global descriptors produced by LPR algorithms in both SLAM and localization tasks is conducted.

\subsection{SLAM}\label{subsec:localization_metrics}

Localization performance is primarily evaluated using the Relative Pose Error (RPE) for local accuracy and Absolute Pose Error (APE) for global consistency \cite{sturm2012benchmark}. As defined in Eq. \ref{eq:rpe_ape}, RPE averages the drift over short intervals, while APE compares the full trajectory against ground truth ($T_{gt}$):

\begin{equation}
    \text{RPE} = \frac{1}{n} \sum_{i=1}^{n} \| (T_{est}^{-1} T_{gt})_i \|, \quad
    \text{APE} = \frac{1}{n} \sum_{i=1}^{n} \| T_{est} - T_{gt} \|
    \label{eq:rpe_ape}
\end{equation}

To provide statistical depth, recent studies \cite{hodson2022root, lodetti2022mae, karunasingha2022root} advocate for the Root Mean Square Error (RMSE), defined as $\text{RMSE} = \sqrt{\frac{1}{n} \sum (y_i - \hat{y}_i)^2}$, which is sensitive to large errors \cite{willmott2005advantages, chai2014root}, and the Mean Absolute Error (MAE = $\frac{1}{n} \sum |y_i - \hat{y}_i|$), which treats errors uniformly.

Geometric accuracy is further dissected into translational and rotational components. The Mean Translation Error (MTE) and L2 norm quantify Euclidean distance deviations ($\| p_{est} - p_{gt} \|$), often reported alongside the Angular Error (AE) to assess orientation drift \cite{pire2017s}. The AE is crucial for visual-inertial systems and is computed as:

\begin{equation}
    \text{AE} = \arccos\left(\frac{tr(R_{est} R_{gt}^T) - 1}{2}\right)
    \label{eq:angular_error}
\end{equation}
Where $R_{est}$ and $R_{gt}$ are the estimated and ground truth rotation matrices, respectively.

\subsection{Localization}\label{subsec:PR_metrics}
The performance of LPR algorithms is typically evaluated using metrics such as precision, recall, and F1-score \cite{terven2025comprehensive}. Precision measures the proportion of correctly recognized places among all recognized places, while recall measures the proportion of correctly recognized places among all actual places. Equation \ref{eq:precision} and Equation \ref{eq:recall} present the formulas for the precision and the recall, respectively.

\begin{equation}
    \text{P} = \frac{TP}{TP + FP}
    \label{eq:precision}
\end{equation}

\begin{equation}
    \text{R} = \frac{TP}{TP + FN}
    \label{eq:recall}
\end{equation}

Where TP is the number of true positives, FP is the number of false positives, and FN is the number of false negatives.

The F1-score is the harmonic mean of precision and recall, providing a single metric that balances both aspects. Another commonly used metric is the area under the precision-recall curve (AUC-PR), which summarizes the trade-off between precision and recall at different thresholds. Additionally, some studies also report the average recall at top-K retrievals, which measures the proportion of correctly recognized places among the top K retrieved places \cite{nouri2025evaluation, schubert2023visual, keetha2023anyloc}. Equation \ref{eq:auc_pr} shows the formula for the AUC-PR.

\begin{equation}
    \text{AUC} = \int_{0}^{1} \text{P}(r) dr
    \label{eq:auc_pr}
\end{equation}
Where P(r) is the precision at recall r.

Finally, both the F1 and F2 scores are also used in some works to evaluate the performance of LPR algorithms. The F1-score gives equal weight to precision and recall, while the F2-score gives more weight to recall. Equation \ref{eq:f1} and Equation \ref{eq:f2} show the formulas for the F1 and F2 scores, respectively.
\begin{equation}
    \text{F1} = 2 \cdot \frac{\text{P} \cdot \text{R}}{\text{P} + \text{R}}
    \label{eq:f1}
\end{equation}
\begin{equation}
    \text{F2} = 5 \cdot \frac{\text{P} \cdot \text{R}}{4 \cdot \text{P} + \text{R}}
    \label{eq:f2}
\end{equation}

Table \ref{tab:agri_lpr_results} provides a concise overview of the outcomes derived from the LPR algorithms, as previously outlined in Table \ref{tab:agri_lpr_methods}, encompassing all the available metrics. It is important to note that, in the case of the method KSI \cite{de2025keypoint}, the results are displayed as a mean of the individual results in the two rows for which the results are provided in the original publication. In the case of TriLoc-NetVLAD \cite{sun2024triloc}, the results are presented separately for same-season and cross-season evaluations. We present the results provided by their novel method which includes Multi-layer Spatial Context based on Cloud Density (MSCD) and the proposed effective layer selection strategy. For SPVSoAP3D \cite{barros2024spvsoap3d}, we present the mean results obtained for the different sequences presented, which range from June to November. These sequences were evaluated individually under their full contribution, which includes the log-euclidian projection and the power normalization steps. Finally, for PointNetPGAP-SLC \cite{barros2024pointnetpgap}, we present the mean results obtained for the different sequences evaluated in their work. As with SPVSoAP3D, these sequences cover the period from June to November, and the results are the mean of the individual results for each sequence.

The results presented in this table suggest that the majority of the methods are evaluated using the same season for which the model was trained. This approach has been demonstrated to yield higher recall values. Conversely, cross-season evaluations frequently yield substantially lower recall values, attributable to the considerable alterations in the environment that occur between seasons. This underscores the complexity inherent in the development of robust place recognition algorithms capable of adapting to the seasonal variations present in agricultural environments. It is also observed that when the season used for both training and testing is data collected on summer, the results are significantly worse than when using data collected on spring. This phenomenon is likely attributable to the fact that, during summer, the vegetation is denser and there are more occlusions present, making it more difficult for the algorithms to recognize places accurately.

\begin{table}[h]
    \centering
    \caption{Summary of the results of the main place recognition methods specifically designed for agricultural environments.}
    \begin{tabular}{|c|c|c|c|c|c|c|}
        \toprule
        \textbf{Type} & \textbf{Method} & \textbf{Year} & \textbf{Dataset} & \textbf{Recall@1} & \textbf{Recall@1\%} & \textbf{MTE [cm]}\\
        \midrule
        \multirow{4}{*}{Vision} & KSI \cite{de2025keypoint} (April) & 2025 & BLT \cite{polvara2024bacchus} & 0.99 & - & 9.22 \\
        & KSI \cite{de2025keypoint} (May) & 2025 & BLT \cite{polvara2024bacchus} & 0.99 & - & 10.905 \\
        & KSI \cite{de2025keypoint} (June) & 2025 & BLT \cite{polvara2024bacchus} & 0.80 & - & 11.83 \\
        & KSI \cite{de2025keypoint} (September) & 2025 & BLT \cite{polvara2024bacchus} & 0.64 & - & 196.62 \\
        \hline
        \multirow{8}{*}{LiDAR} & ORCHNet \cite{barros2023orchnet} (PointNet) & 2023 & HORTO-3DLM \cite{barros2024spvsoap3d} & 0.52 & 0.92 & - \\
        & ORCHNet \cite{barros2023orchnet} (ResNet) & 2023 & HORTO-3DLM \cite{barros2024spvsoap3d} & 0.45 & 0.94 & - \\
        & TriLoc-NetVLAD (same season)\cite{sun2024triloc} & 2024 & Vineyard (inhouse) & 0.75 & - & -\\
        & TriLoc-NetVLAD (cross season)\cite{sun2024triloc} & 2024 & Vineyard (inhouse) & 0.55 & - & -\\
        & SPVSoAP3D \cite{barros2024spvsoap3d} & 2024 & HORTO-3DLM \cite{barros2024spvsoap3d} & 0.63 & 0.96 & -\\
        & PointNetPGAP-SLC \cite{barros2024pointnetpgap} & 2024 & HORTO-3DLM \cite{barros2024spvsoap3d} & 0.75 & - & -\\
        & MinkUNeXt-VINE (same season)\cite{vilellacantos2026lowcosthighefficiency} & 2026 & TEMPO-VINE \cite{martini2025tempo} & 0.68 & 0.97 & -\\
        & MinkUNeXt-VINE (cross season)\cite{vilellacantos2026lowcosthighefficiency} & 2026 & TEMPO-VINE \cite{martini2025tempo} & 0.41 & 0.70 & -\\
        \bottomrule
    \end{tabular}
    \label{tab:agri_lpr_results}
\end{table}

\subsection{Clustering}\label{subsec:clustering_metrics}
Global descriptors generated by LPR algorithms can be used for clustering. The performance  on this application is typically evaluated using metrics such as the silhouette score \cite{rousseeuw1987, shahapure2020} and the Davies-Bouldin index \cite{davies1979}. The silhouette score measures the similarity of each point to its own cluster compared to other clusters, while the Davies-Bouldin index measures the average similarity between each cluster and its most similar cluster. Equation \ref{eq:silhouette} shows the formula for the silhouette score, while Equation \ref{eq:davies_bouldin} shows the formula for the Davies-Bouldin index.

\begin{equation}
    \text{s(i)} = \frac{b(i) - a(i)}{\max(a(i), b(i))}
    \label{eq:silhouette}
\end{equation}
Where $a(i)$ is the average distance between point i and all other points in its own cluster, and $b(i)$ is the average distance between point i and all other points in the nearest cluster.

\begin{equation}
    \text{DB} = \frac{1}{k} \sum_{i=1}^{k} \max_{j \neq i} \left( \frac{s_i + s_j}{d_{ij}} \right)
    \label{eq:davies_bouldin}
\end{equation}

Where $k$ is the number of clusters, $s_i$ is the average distance between points in cluster i, $s_j$ is the average distance between points in cluster j, and $d_{ij}$ is the distance between the centroids of clusters i and j.

To provide a more rigorous evaluation, several works utilize the Adjusted Rand Index (ARI) \cite{santos2009use}. The ARI compensates for the expected overlap due to pure chance, providing a more reliable indicator of clustering quality as expressed in Equation \ref{eq:ari}.

\begin{equation}
    \text{ARI} = \frac{\sum_{ij} \binom{n_{ij}}{2} - \left[ \sum_i \binom{a_i}{2} \sum_j \binom{b_j}{2} \right] / \binom{n}{2}}{\frac{1}{2} \left[ \sum_i \binom{a_i}{2} + \sum_j \binom{b_j}{2} \right] - \left[ \sum_i \binom{a_i}{2} \sum_j \binom{b_j}{2} \right] / \binom{n}{2}}
    \label{eq:ari}
\end{equation}
Where $n_{ij}$ is the number of points in both cluster i and cluster j, $a_i$ is the number of points in cluster i, $b_j$ is the number of points in cluster j, and n is the total number of points.

\section{Conclusions}\label{sec:conclusions}
The achievement of robust LPR stands as a critical milestone for the operational success of autonomous mobile robotics. Unlike static urban scenarios, agricultural environments present unique challenges—such as unstructured terrain and severe seasonal variability—that render standard urban-centric solutions ineffective. Consequently, developing specialized LPR approaches is not merely a technical necessity but a strategic one. Reliable localization in GNSS-denied environments is the key enabler for precision agriculture, allowing for optimized food production and mitigating the impact of the growing global agricultural workforce shortage.

To support this goal, this work presented a comprehensive review of the SOTA in agricultural localization, focusing on general agricultural DL literature from 2020 onwards and the existing specific LPR works in these settings. Moreover, we provided an analysis of the current results found in the scarce agricultural LPR literature. Although there have been improvements in recent years, cross-season validation remains a significant challenge, with recall values often dropping below 50\%. We also observed significant performance differences between winter and summer campaigns, primarily due to severe environmental occlusions, with a difference in the Recall@1 metric of more than 20\%. These findings underscore the necessity for additional research to develop robust algorithms that can handle the dynamic nature of agricultural environments.

We discussed the distinct environmental constraints and provided a critical analysis of current datasets and metrics. Notably, our comparison highlights a disparity between the abundance of computer vision datasets and the scarcity of data specifically curated for localization tasks. It is our hope that this survey clarifies the current landscape and guides future research toward bridging the gap between theoretical algorithms and real-world agricultural autonomy.

\backmatter

\bmhead{Acknowledgements}

This research work is part of the project PID2023-149575OB-I00 funded by MICIU/AEI/10.13039/501100011033 and by FEDER, UE. It is also part of the project CIPROM/2024/8, funded by Generalitat Valenciana, Conselleria de Educación, Cultura, Universidades y Empleo (program PROMETEO 2025).

\bmhead{Author contributions}
Judith Vilella-Cantos: Conceptualization, Methodology, Formal analysis, Investigation, Writing—original draft preparation, Writing—review and editing. Mónica Ballesta: Resources, Writing—review and editing, Supervision, Formal analysis, Visualization. David Valiente: Resources, Writing—review and editing, Supervision, Formal analysis, Visualization. María Flores: Resources, Visualization, Funding acquisition. Luis Payá: Project administration, Funding acquisition, Visualization.

\bmhead{Competing interests}
The authors declare that they have no competing interests.

\bibliography{bibliography}

\end{document}